\documentclass[letterpaper, 10 pt, conference]{ieeeconf}  

\IEEEoverridecommandlockouts                              

\usepackage{graphics} 
\usepackage{amsmath} 
\usepackage{amssymb}  
\usepackage{tikz}
\usetikzlibrary{arrows.meta,calc,angles,quotes}
\usepackage{newtxtext,newtxmath}
\usepackage{booktabs}
\usepackage{siunitx}
\usepackage{cleveref}

\title{\LARGE \bf
Radar-Native Control Barrier Functions: Doppler Range Rate as a Measured Barrier Derivative}

\author{Johannes Mootz$^{1,2}$, Hesam Mojtahedi$^{1,3}$ and Reza Akhavian$^{1}$%
\thanks{$^{1}$Department of Civil, Construction, and Environmental Engineering,
        San Diego State University, San Diego, CA 92182, USA.
        {\tt\small \{jmootz, hmojtahedi, rakhavian\}@sdsu.edu}}%
\thanks{$^{2}$Department of Mechanical and Aerospace Engineering,
        University of California San Diego, La Jolla, CA 92093, USA.}%
\thanks{$^{3}$Department of Electrical and Computer Engineering,
        University of California San Diego, La Jolla, CA 92093, USA.}%
}

\begin{document}

\maketitle
\thispagestyle{empty}
\pagestyle{empty}

\begin{abstract}

Safe control around dynamic obstacles requires information about both distance and relative motion. Range-based filters commonly reconstruct the required range rate from successive range measurements or from obstacle-state estimators, which can introduce estimator delay or sensitivity to range noise when obstacle motion changes.
A frequency-modulated continuous-wave (FMCW) radar measures radial relative velocity directly through Doppler. We use this measurement as the derivative of an ego-centered range control barrier function (CBF) and obtain a closed-form filter on the robot's forward velocity.
A measurement margin constructed from sensor error bounds accounts for errors in range, Doppler, ego velocity, and bearing. In 508 paired encounters within the platform's retreat capability, direct Doppler produces two buffer incursions, compared with 15 for a constant-acceleration and 25 for a constant-velocity Kalman filter.
On a Unitree Go1, the controller stops at $0.777\pm0.015\,\mathrm{m}$ against the predicted $0.758\,\mathrm{m}$ and reacts to a farther approaching object while a nearer static return remains. 
The guarantee covers the longitudinal range channel under bounded measurement errors and continuously available detections.
Direct Doppler removes the need to reconstruct range rate in the safety loop, reducing the delay between a change in obstacle motion and its effect on the safety command.

\end{abstract}

\section{INTRODUCTION}

When a dynamic obstacle starts moving toward a robot, the safety command must respond before the remaining distance becomes too small. This requires information about both distance and relative motion. 
The nearest obstacle is not necessarily the one that requires the strongest response: a farther approaching object can require braking or retreat while a nearer object remains stationary. 
Control barrier functions (CBFs) provide a way to enforce a distance constraint by filtering the robot's nominal command under specified conditions \cite{ames2019control}.

For a range-based barrier, the constraint depends on the range rate. Range-only measurements, such as from conventional LiDAR or depth cameras, do not provide this quantity directly, so it must be reconstructed from successive measurements or obtained from an obstacle-state estimator \cite{long2026sensor}. 
When obstacle motion changes, the estimate can take several samples to adjust. Increasing the estimator's responsiveness also passes more range noise into the control command. 
For a safety filter, this tradeoff matters most when little clearance remains and the constraint becomes active.

A frequency-modulated continuous-wave (FMCW) radar provides range and bearing information together with radial velocity measured through the Doppler effect \cite{bilik2019rise}. 
Each detection carries its radial velocity relative to the sensor, obtained within a single frame. For a range barrier, this is the barrier derivative. 
The range rate is therefore available without differentiating successive range measurements or estimating an obstacle-velocity state.
To evaluate a candidate velocity command, the contribution of the robot's current forward motion must first be accounted for.

We present a radar-native CBF formulation that uses this measured range rate and filters the forward velocity command. The relative range dynamics give the correction from the measured ego velocity to a candidate command. 
Each retained detection then imposes an upper velocity bound, and the smallest bound determines the safety command. A measurement-error margin accounts for bounded errors in range, Doppler, ego velocity, and bearing. It follows from the sensor configuration and can be checked before deployment, yet still has to be accounted for in the constraint.
The formulation acts on the velocity interface available on many robots \cite{singletary2021safety} and covers the longitudinal range channel under the conditions stated in Section~\ref{sec:robust_constraint}.

The evaluation examines how the source of range rate affects the safety command after a change in obstacle motion. In 508 paired encounters within the platform's retreat capability, direct Doppler produces two buffer incursions, compared with 15 for a constant-acceleration and 25 for a constant-velocity Kalman filter. 
Hardware tests on a Unitree Go1 examine the predicted static standoff and the response to a farther approaching object while a nearer static return remains. The main contributions are:

\begin{itemize}
\item A radar-native range CBF that uses FMCW Doppler as its measured derivative and yields a closed-form filter for forward velocity, without an obstacle-velocity estimator.
\item A measurement-error margin expressed in radar and ego-velocity quantities, with explicit conditions for forward invariance.
\item A paired simulation showing fewer near-boundary incursions than range-based Kalman filters, and hardware tests of predicted standoff and multi-object response on a quadruped.
\end{itemize}

\section{RELATED WORK}

\subsection{CBFs for Collision Avoidance}
Control barrier functions are widely used for safety-critical control and can ensure the forward invariance of a safe set \cite{ames2019control}. 
Applications include adaptive cruise control \cite{ames2014control} and collision avoidance using collision-cone barriers for moving obstacles \cite{tayal2026collision}.

The chosen barrier and robot model determine how the control input enters the constraint. Higher-order formulations address constraints whose derivatives do not directly depend on the available input. 
However, they may require acceleration- or torque-level control, which is often hidden within internal control loops when robots expose only first-order velocity commands.
On robots that expose only velocity commands, safety can be enforced on a reduced-order velocity model \cite{singletary2021safety, molnar2021model}. The guarantee then transfers to the physical robot under a condition on the velocity-tracking error.

\subsection{Safety from Sensor Measurements}
A common problem in applying CBFs is the perception gap between theory and the limits of real sensors on robots. While many formulations assume full state information about the robot and its environment, this assumption is rarely met in real-world environments. Existing approaches address this gap with robustness margins, such as in measurement-robust CBFs (MR-CBFs) \cite{dean2021guaranteeing, cosner2021measurement}.
Other routes include conformal prediction \cite{yang2023safe}, distributionally robust sensor-native \cite{long2026sensor}, and observer-based \cite{wang2022observer} formulations. Sampled-data formulations account for changes between control updates \cite{oruganti2023robust}.

The required state information is often obtained from external motion capture or other globally referenced sources, and more recently from onboard LiDAR. LiDAR can close part of the perception gap by constructing a CBF directly from point-cloud geometry \cite{dai2024sailing} or by building a smooth safety function from the accumulated measurements, for example a Gaussian-process field \cite{keyumarsi2023lidar} or a signed-distance function derived from an occupancy grid \cite{raja2024ogm}.

For dynamic obstacles, the constraint needs information about relative motion. A common assumption is that the velocities or accelerations of unknown obstacles are bounded \cite{long2026sensor, zhang2026collision}. This is accepted engineering practice, but the resulting guarantees are conditional on the bounds holding during operation. Long et al.\ \cite{long2026sensor} also discuss obtaining obstacle-motion derivatives from radar, Doppler LiDAR, or a velocity-estimation algorithm within their distributionally robust formulation.

Measured output derivatives are used in \cite{autenrieb2025sensor} in an incremental CBF formulation. Their reduced-order model follows from a local Taylor approximation of the output dynamics, with margins for approximation and measurement errors. 
In our formulation, the correction for a candidate forward-velocity command follows exactly from the relative range dynamics at the measurement instant under the velocity-input model.

\subsection{Radar-Based Control}

Doppler range rate is well established in automotive radar \cite{bilik2019rise}, while its use in robotics remains less explored \cite{harlow2024new}.
Its lower angular resolution and artifacts such as ghost detections complicate its use for safety-critical control \cite{bruhl2024making}.
However, radar can support perception under adverse weather conditions in which optical measurements degrade \cite{bijelic2020seeing}. The single-chip sensors used in robotics also typically deliver fewer points per frame than a scanning LiDAR.

Using Doppler returns directly, without an object-level tracker, is established for radar ego-velocity estimation in visually degraded environments \cite{kramer2020radar}.
Reactive filters that work from local range and bearing measurements operate at the same command interface \cite{zhang2026collision}, but they reconstruct the barrier derivative from geometry and ego motion and therefore still require an assumption on the obstacle velocity.

Existing radar obstacle-avoidance systems use Doppler for detection and gating, and then pass tracked obstacle states to the controller. One example combines a radial-velocity gate, DBSCAN clustering, and an interacting-multiple-model tracker with a distributionally robust acceleration-level CBF \cite{mandaokar2026radar}.
Doppler is used there to select dynamic points before clustering, and the barrier constraint is evaluated using the tracked obstacle states. The controller accounts for uncertainty in these states and acts on acceleration.
The formulation here uses the Doppler range rate and assumed measurement-error bounds in a constraint on the first-order velocity command that many robots expose.

\section{PROBLEM FORMULATION}

Let the robot position be \(\mathbf{p}=[p_x,p_y]^\top\) and its heading be \(\theta\). The robot is represented by the reduced-order unicycle kinematics
$\dot{\mathbf{p}}=v[\cos\theta,\sin\theta]^\top$ and
$\dot{\theta}=\omega$, where $v$ is the forward velocity and $\omega$ is the yaw rate. We use this reduced-order model at the command level, following the idea in \cite{singletary2021safety}.

Consider a point obstacle $i$ in an ego-centered polar coordinate system in which $R_i$ is the distance to the robot and $\gamma_i$ is the bearing relative to the robot heading.

Let $v_{o,r,i}$ and $v_{o,t,i}$ denote the components of the obstacle velocity along and perpendicular to the line of sight. The relative polar dynamics are

\begin{equation}
\begin{aligned}
\begin{bmatrix}
    \dot{R}_i \\
    \dot{\gamma}_i
\end{bmatrix}
&=
\underbrace{
\begin{bmatrix}
    v_{o,r,i} \\
    v_{o,t,i}/R_i
\end{bmatrix}
}_{\text{obstacle-motion drift}} +
\underbrace{
\begin{bmatrix}
    -\cos(\gamma_i) & 0 \\
    \sin(\gamma_i)/R_i & -1
\end{bmatrix}
}_{\text{control matrix}}
\begin{bmatrix}
    v \\
    \omega
\end{bmatrix}.
\end{aligned}
\label{eq:relative_polar_dynamics}
\end{equation}

For a stationary obstacle, these equations reduce to the standard polar-coordinate unicycle model in \cite{todorovski2026modular}. The singularity at \(R_i=0\) lies outside the safety domain considered below because \(R_{\mathrm{safe}}>0\).

For each retained radar point $i$, the available measurement is
\begin{equation}
\hat{\mathbf{z}}_i
=
\begin{bmatrix}
    \hat{R}_i &
    \hat{\gamma}_i &
    \widehat{\dot{R}}_i
\end{bmatrix}^{\top},
\label{eq:radar_measurement}
\end{equation}
where $\hat{R}_i$, $\hat{\gamma}_i$, and $\widehat{\dot{R}}_i$ are the measured range, bearing, and Doppler range rate, respectively. Here, $\widehat{\dot{R}}_i$ is measured directly through the Doppler effect and is not obtained by differentiating consecutive range measurements. A negative range rate, $\widehat{\dot{R}}_i<0$, indicates that the obstacle and robot are closing.

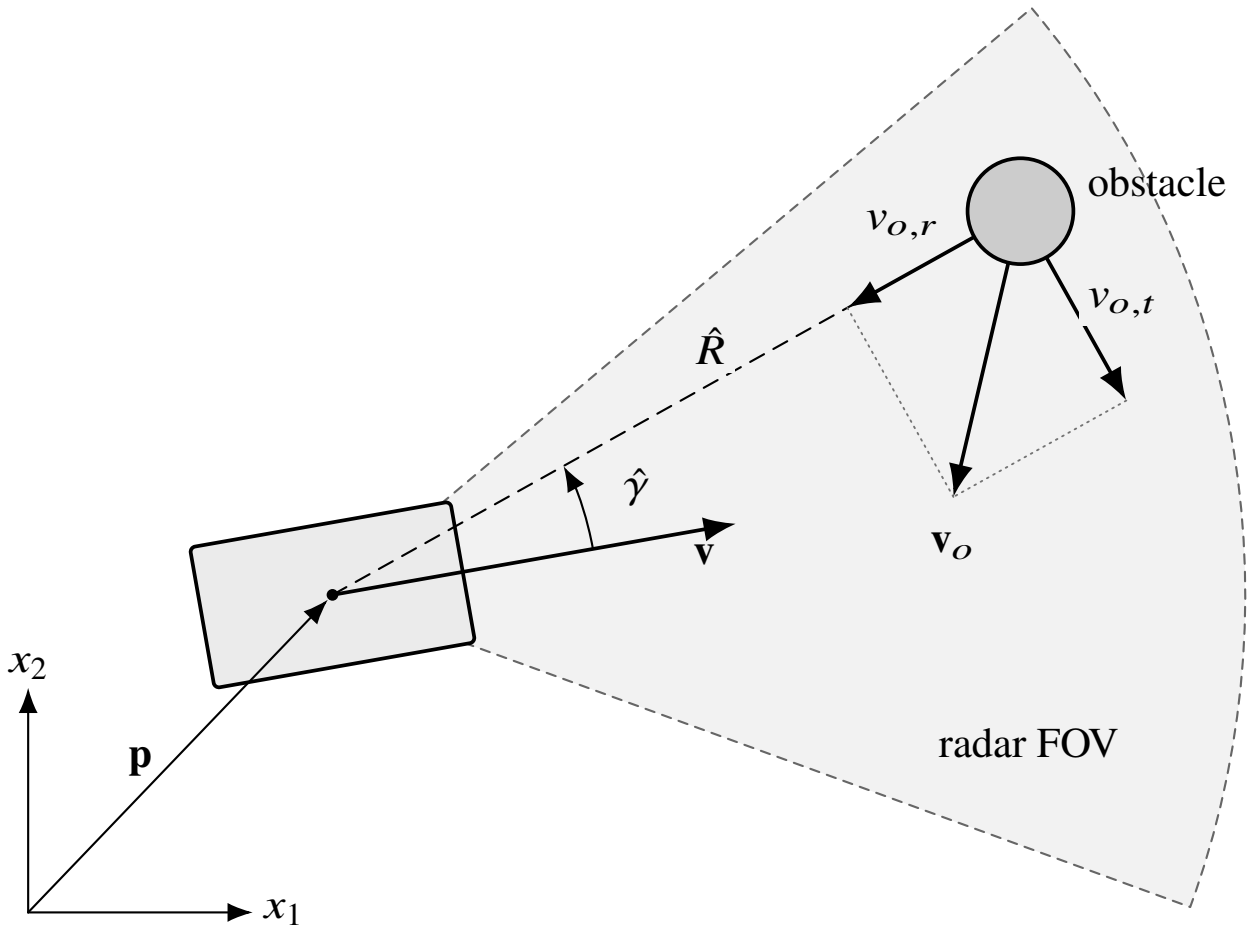
\begin{figure}[t]
\centering
\begin{tikzpicture}[
    scale=1.75,                     
    >=Latex,
    line cap=round, line join=round,
    every node/.style = {font=\footnotesize, inner sep=1.5pt},
    axis/.style       = {->, line width=0.4pt},
    vec/.style        = {->, line width=0.7pt},
    posvec/.style     = {->, line width=0.4pt},
    dashedline/.style = {densely dashed, line width=0.4pt},
    fovline/.style    = {dash pattern=on 1.8pt off 1.5pt, line width=0.4pt, black!60},
    guide/.style      = {densely dotted, line width=0.35pt, black!55},
    dot/.style        = {circle, fill=black, inner sep=0pt, minimum size=2.2pt},
    lbl/.style        = {fill=fovfill, inner sep=1.5pt}
]

\pgfmathsetmacro{\hdg}{10}          
\pgfmathsetmacro{\fovhalf}{30}      
\pgfmathsetmacro{\fovr}{3.45}       
\colorlet{fovfill}{black!5}         

\coordinate (A)   at (0.30,0.35);   
\coordinate (E)   at (1.45,1.55);   
\coordinate (Obs) at (4.05,3.00);   

\pgfmathanglebetweenpoints
    {\pgfpointanchor{E}{center}}
    {\pgfpointanchor{Obs}{center}}
\pgfmathsetmacro{\losang}{\pgfmathresult}
\pgfmathsetmacro{\radang}{\losang+180}   
\pgfmathsetmacro{\tanang}{\losang-90}    

\pgfmathanglebetweenpoints
    {\pgfpointanchor{A}{center}}
    {\pgfpointanchor{E}{center}}
\pgfmathsetmacro{\posang}{\pgfmathresult}

\fill[fovfill]
    (E) -- ++(\hdg-\fovhalf:\fovr)
    arc[start angle=\hdg-\fovhalf, end angle=\hdg+\fovhalf, radius=\fovr]
    -- cycle;

\draw[fovline]
    (E) -- ++(\hdg-\fovhalf:\fovr)
    arc[start angle=\hdg-\fovhalf, end angle=\hdg+\fovhalf, radius=\fovr]
    -- cycle;

\node at ($(E)+(\hdg-\fovhalf+8:0.78*\fovr)$) {radar FOV};

\draw[axis] (A) -- ++(0.85,0) node[right] {$x_1$};
\draw[axis] (A) -- ++(0,0.85) node[above] {$x_2$};

\begin{scope}[shift={(E)}, rotate=\hdg]
    \draw[line width=0.7pt, rounded corners=1pt, fill=black!8]
        (-0.50,-0.27) rectangle (0.50,0.27);
    \draw[vec] (0,0) -- (1.55,0) node[pos=0.92, below=1pt] {$\mathbf v$};
\end{scope}

\node[dot] at (E) {};               
\draw[posvec, shorten >=1.2pt] (A) -- (E)
    node[pos=0.42, above left=-1pt] {$\mathbf p$};

\draw[dashedline] (E) -- (Obs);
\node[lbl, above=2pt] at ($(E)!0.55!(Obs)$) {$\hat{R}$};

\coordinate (HeadingRef) at ($(E)+(\hdg:2.0)$);
\coordinate (LOSRef)     at ($(E)+(\losang:2.0)$);

\pic[draw, ->, line width=0.4pt,
     angle radius=1.75cm, angle eccentricity=1.22, "$\hat{\gamma}$"]
    {angle = HeadingRef--E--LOSRef};


\coordinate (VrEnd) at ($(Obs)+(\radang:0.75)$);
\coordinate (VtEnd) at ($(Obs)+(\tanang:0.82)$);
\coordinate (VoEnd) at ($(VrEnd)+(VtEnd)-(Obs)$);

\draw[guide] (VrEnd) -- (VoEnd);
\draw[guide] (VtEnd) -- (VoEnd);

\draw[vec] (Obs) -- (VrEnd);
\draw[vec] (Obs) -- (VtEnd);
\draw[vec] (Obs) -- (VoEnd);

\filldraw[fill=black!20, line width=0.7pt] (Obs) circle (0.20);
\node[anchor=south west] at ($(Obs)+(0.22,0.02)$) {obstacle};

\node[lbl] at ($(Obs)!0.55!(VrEnd)+(\losang+90:0.17)$) {$v_{o,r}$};
\node[lbl] at ($(Obs)!0.60!(VtEnd)+(\losang:0.17)$)    {$v_{o,t}$};
\node[lbl, anchor=north] at ($(VoEnd)+(0,-0.10)$)      {$\mathbf v_o$};

\end{tikzpicture}

\caption{Planar radar measurement model. A return from the obstacle is observed at range $\hat{R}$ and bearing $\hat{\gamma}$ relative to the robot's heading. The obstacle velocity $\mathbf v_o$ is decomposed into radial and tangential components $v_{o,r}$ and $v_{o,t}$.}
\label{fig:polar_model}
\end{figure}

The Doppler range rate combines obstacle radial motion and ego motion according to the first row of \eqref{eq:relative_polar_dynamics}:
$\dot{R}_i=v_{o,r,i}-v\cos(\gamma_i)$.
Fig.~\ref{fig:polar_model} illustrates the model and the corresponding sign conventions. The radar returns are three dimensional in general and a planar case is assumed here.

Given nominal commands \(v_{\mathrm{nom}}\) and \(\omega_{\mathrm{nom}}\), the radar measurements, and an ego-velocity estimate \(\hat v\), the task is to determine a forward-velocity command \(v_{\mathrm{cmd}}\in[v_{\min},v_{\max}]\) that maintains \(R_i(t)\geq R_{\mathrm{safe}}\) for the considered obstacle points while limiting changes to \(v_{\mathrm{nom}}\). The yaw-rate command remains \(\omega_{\mathrm{nom}}\). We consider initially safe configurations, bounded measurement errors, and intervals over which the obstacle points remain available and the required clearance can be maintained within the velocity limits.

\section{METHODOLOGY}

\subsection{Control Barrier Function Preliminaries}
 
We recall CBFs from \cite{ames2019control} and consider the control-affine system

\begin{equation}
\dot{x}=f(x)+g(x)u,
\label{eq:control_affine_system}
\end{equation}

where \(x\in\mathcal{D}\subseteq\mathbb{R}^{n}\) is the system state, \(u\in\mathcal{U}\subseteq\mathbb{R}^{m}\) is the control input, and \(f\) and \(g\) are locally Lipschitz.
Let \(h:\mathcal{D}\rightarrow\mathbb{R}\) be continuously differentiable and define the safe set \(\mathcal{C}=\{x\in\mathcal{D}\mid h(x)\geq0\}\).
The derivative along the system trajectories is $
\dot{h}(x,u)=L_fh(x)+L_gh(x)u$.

The function \(h\) is a CBF on \(\mathcal{D}\) if there exists an extended class-\(\mathcal{K}_{\infty}\) function \(\alpha_h\) such that
\begin{equation}
\sup_{u\in\mathcal{U}}
\left[
L_fh(x)+L_gh(x)u+\alpha_h(h(x))
\right]
\geq0,
\quad \forall x\in\mathcal{D}.
\label{eq:cbf_condition}
\end{equation}

If \(\nabla h(x)\neq0\) for all \(x\in\partial\mathcal{C}\) and a locally Lipschitz feedback controller \(u=k(x)\in\mathcal{K}_{\mathrm{CBF}}(x)\) exists, then \(\mathcal{C}\) is forward invariant over the maximal interval of existence of the closed-loop solution \cite{ames2019control}. If the closed-loop system is forward complete, \(x(0)\in\mathcal{C}\) implies \(x(t)\in\mathcal{C}\) for all \(t\geq0\).

For the linear choice \(\alpha_h(h)=\alpha h\), where \(\alpha>0\), the imposed condition becomes

\begin{equation}
\dot{h}+\alpha h\geq 0.
\label{eq:linear_cbf_constraint}
\end{equation}

For a collection of \(N\) obstacle constraints, the overall safe set is
\begin{equation}
\mathcal{C}
=
\bigcap_{i=1}^{N}\mathcal{C}_i,
\qquad
\mathcal{C}_i
=
\{x\in\mathcal{D}\mid h_i(x)\geq0\}.
\label{eq:multiple_safe_sets}
\end{equation}

Forward invariance of the intersection requires a common admissible input that satisfies \eqref{eq:linear_cbf_constraint} for every \(h_i\).
For moving obstacles, the barrier function is time-varying or defined using relative coordinates, and its derivative depends on relative motion. The proposed formulation exploits radar Doppler measurements to obtain this derivative directly in sensor coordinates.

\subsection{Doppler as the Barrier Derivative}
\label{sec:doppler_derivative}

Equation~\eqref{eq:relative_polar_dynamics} shows that the obstacle motion appears as an unknown drift term. The forward velocity $v$ affects both range and bearing, whereas the yaw rate $\omega$ does not affect the instantaneous range. Consequently, the range-based barrier constrains the forward velocity directly, while steering affects the range only indirectly through \(\gamma_i\).

For each obstacle $i$, we use the range barrier
\begin{equation}
    h_i(R_i)=R_i-R_{\mathrm{safe}},
    \quad
    \dot{h}_i=\dot{R}_i=v_{o,r,i}-v\cos(\gamma_i).
    \label{eq:range_barrier}
\end{equation}
Thus, the radar Doppler measurement $\widehat{\dot{R}}_i$ directly provides the measured barrier derivative, $\widehat{\dot{h}}_i=\widehat{\dot{R}}_i$.
The Doppler range rate contains the effect of the robot's current ego motion, represented by the measured ego velocity $\hat{v}$. 

At the current measurement instant, the range rate corresponding to a candidate command \(v_{\mathrm{cmd}}\) is therefore
\begin{equation}
    \widehat{\dot{R}}_i(v_{\mathrm{cmd}})
    =
    \widehat{\dot{R}}_i
    +
    (\hat{v}-v_{\mathrm{cmd}})\cos(\hat{\gamma}_i).
    \label{eq:command_adjusted_range_rate}
\end{equation}
This adjustment changes only the forward ego-velocity contribution. The obstacle motion and the remaining components of the sensor motion are contained in the measured range rate and are held fixed during the instantaneous command evaluation.

Substituting \eqref{eq:range_barrier} and \eqref{eq:command_adjusted_range_rate} into \eqref{eq:linear_cbf_constraint} gives
\begin{equation}
\begin{aligned}
    &\widehat{\dot{R}}_i
    +(\hat{v}-v_{\mathrm{cmd}})\cos(\hat{\gamma}_i)
    +\alpha(\hat{R}_i-R_{\mathrm{safe}})
    \geq 0.
\end{aligned}
    \label{eq:doppler_cbf_constraint}
\end{equation}

The forward corridor is selected such that $\cos(\hat{\gamma}_i)>0$. Each retained detection then imposes the upper velocity bound
\begin{equation}
    \begin{aligned}
        v_{\mathrm{cmd}} &\leq v_{\mathrm{cap},i} := \hat{v}
        + \frac{
            \widehat{\dot{R}}_i
            + \alpha(\hat{R}_i - R_{\mathrm{safe}})
        }{
            \cos(\hat{\gamma}_i)
        }.
    \end{aligned}
    \label{eq:velocity_cap}
\end{equation}

Let $\mathcal{I}$ denote the set of retained detections and let the nominal velocity command satisfy \(v_{\mathrm{nom}}\in[v_{\min},v_{\max}]\), with \(v_{\max}=v_{\mathrm{des}}\) and \(v_{\min}=-v_{\mathrm{back}}\). The safety filter is

\begin{equation}
    v_{\mathrm{cmd}}
    =
    \max\Big\{v_{\min},\;
    \min\{v_{\mathrm{nom}},
    \min_{i\in\mathcal{I}}v_{\mathrm{cap},i}\}\Big\}.
    \label{eq:scalar_safety_filter}
\end{equation}
If $\mathcal{I}$ is empty, then $v_{\mathrm{cmd}}=v_{\mathrm{nom}}$. Each active detection imposes an upper bound on the same scalar velocity command, and the most restrictive detection determines the safety-imposed upper bound. 
Since every active constraint is an upper bound on the same scalar input, the filter has a closed-form solution and does not require a QP solver.

The constraints are feasible when \(v_{\mathrm{cap},i}\geq v_{\min}\) for every \(i\in\mathcal I\). The outer maximum is then inactive and every constraint is satisfied. Otherwise no admissible forward-velocity command satisfies all active constraints, the outer maximum is active, and the filter commands \(v_{\min}\) and the frame is infeasible. The upper limit needs no bound because \(v_{\mathrm{nom}}\leq v_{\max}\).

The yaw-rate command remains unchanged, \(\omega_{\mathrm{cmd}}=\omega_{\mathrm{nom}}\). Under the reduced-order model, the yaw rate does not enter the instantaneous range dynamics, and the proposed filter therefore certifies only the longitudinal range channel.

Under exact measurements, exact tracking of the velocity command, and continuous enforcement of the feasible constraint, each active range barrier is forward invariant. Proposition~1 in Section~\ref{sec:robust_constraint} states this together with the measurement margin. If \(h_i<0\), the command may promote recovery, but forward invariance no longer applies.

\subsection{Measurement-Robust Constraint}
\label{sec:robust_constraint}

The nominal formulation assumes exact measurements. Following measurement-robust CBF formulations \cite{dean2021guaranteeing, cosner2021measurement}, we tighten the constraint by the worst-case effect of the measurement error. Assume bounded errors on the measured quantities $|\hat R_i-R_i|\leq\bar e_R$, $|\widehat{\dot R}_i-\dot R_i|\leq\bar e_{\dot R}$, $|\hat v-v|\leq\bar e_v$ and $|\hat\gamma_i-\gamma_i|\leq\bar e_\gamma$.

Let \(\Delta v_{\max}\) bound the commanded velocity change, such that \(|\hat v-v_{\mathrm{cmd}}|\leq\Delta v_{\max}\). The difference between the measured and the true left-hand side of \eqref{eq:doppler_cbf_constraint} is then bounded by
\begin{equation}
\varepsilon_{\mathrm{meas}}
=
\underbrace{
\bar e_{\dot R}
+
\bar e_v
+
(\Delta v_{\max}+\bar e_v)\,\bar e_\gamma
}_{\varepsilon_{\mathrm{rate}}}
+
\alpha\bar e_R .
\label{eq:eps_meas}
\end{equation}
The three terms grouped as \(\varepsilon_{\mathrm{rate}}\) all perturb the same measured range rate. The range term is kept separate because it enters through \(\alpha\).
The factor \(\Delta v_{\max}\) appears because bearing error changes the
line-of-sight projection of the difference between measured and commanded
forward velocity.

The nominal condition is tightened to
\begin{equation}
\widehat{\dot R}_i
+
(\hat v-v_{\mathrm{cmd}})\cos(\hat\gamma_i)
+
\alpha(\hat R_i-R_{\mathrm{safe}})
\geq
\varepsilon_{\mathrm{meas}}.
\label{eq:robust_cbf_constraint}
\end{equation}

The tightened condition replaces the cap in \eqref{eq:velocity_cap} with
\begin{equation}
    v_{\mathrm{cmd}} \leq v_{\mathrm{cap},i} := \hat{v}
    + \frac{\widehat{\dot{R}}_i + \alpha(\hat{R}_i - R_{\mathrm{safe}})
    - \varepsilon_{\mathrm{meas}}}{\cos(\hat{\gamma}_i)},
    \label{eq:cap_robust}
\end{equation} 
and \eqref{eq:scalar_safety_filter} is applied to these caps. The margin reduces the velocity cap by \(\varepsilon_{\mathrm{meas}}/\cos\hat\gamma_i\), so its effect on the allowed forward velocity increases with bearing. \\

\noindent\textbf{Proposition 1 (Measurement-robust range safety)}
Consider an interval with a fixed set $\mathcal I$ of continuously available obstacle points, with $R_i(0)\geq R_{\mathrm{safe}}$ for every $i\in\mathcal I$. Let the error bounds used in \eqref{eq:eps_meas} hold at every time, with $\cos\hat\gamma_i>0$, and $\lvert\hat v-v_{\mathrm{cmd}}\rvert\leq\Delta v_{\max}$. If the command is feasible, is tracked exactly by the continuous-time velocity-input model, and enforces \eqref{eq:robust_cbf_constraint} for every $i\in\mathcal I$, then $R_i(t)\geq R_{\mathrm{safe}}$ throughout the interval.

\emph{Proof.}
For a candidate command, define the measured and true constraint values
\begin{align*}
\hat\ell_i &:= \widehat{\dot R}_i+(\hat v-v_{\mathrm{cmd}})\cos\hat\gamma_i
              +\alpha(\hat R_i-R_{\mathrm{safe}}),\\
\ell_i &:= \dot R_i+(v-v_{\mathrm{cmd}})\cos\gamma_i
              +\alpha(R_i-R_{\mathrm{safe}}).
\end{align*}

The ego-velocity terms split as
$(\hat v-v_{\mathrm{cmd}})\cos\hat\gamma_i-(v-v_{\mathrm{cmd}})\cos\gamma_i
=(\hat v-v)\cos\hat\gamma_i+(v-v_{\mathrm{cmd}})(\cos\hat\gamma_i-\cos\gamma_i)$.
Using $\lvert\cos\hat\gamma_i\rvert\leq1$, $\lvert\cos a-\cos b\rvert\leq\lvert a-b\rvert$, and $\lvert v-v_{\mathrm{cmd}}\rvert\leq\Delta v_{\max}+\bar e_v$,
\[
\lvert\hat\ell_i-\ell_i\rvert
\leq \bar e_{\dot R}+\bar e_v+(\Delta v_{\max}+\bar e_v)\bar e_\gamma
      +\alpha\bar e_R
=\varepsilon_{\mathrm{meas}}.
\]
Equation~\eqref{eq:robust_cbf_constraint} gives $\hat\ell_i\geq\varepsilon_{\mathrm{meas}}$, hence $\ell_i\geq0$. By the first row of \eqref{eq:relative_polar_dynamics}, $\ell_i=v_{o,r,i}-v_{\mathrm{cmd}}\cos\gamma_i+\alpha h_i$, which is $\dot h_i+\alpha h_i$ at $v=v_{\mathrm{cmd}}$. Under exact tracking the robot moves with $v=v_{\mathrm{cmd}}$, so
\[
\frac{d}{dt}\!\left(e^{\alpha t}h_i(t)\right)
=e^{\alpha t}\left(\dot h_i+\alpha h_i\right)\geq0,
\]
so $h_i(t)\geq e^{-\alpha t}h_i(0)\geq0$. 

The argument holds for every retained detection, which proves invariance of their intersection. \hfill\(\square\)

Proposition 1 uses the Doppler measurement to account for instantaneous radial motion. The measurement-error bounds are assumptions that must hold during operation.
Where obstacle-velocity or acceleration bounds are hypotheses about previously unseen objects that cannot be checked while the robot is running, $\bar{e}_{\dot{R}}$, $\bar{e}_R$, $\bar{e}_v$, and $\bar{e}_\gamma$ are properties of the deployed radar and the body-velocity estimate, fixed by the Doppler and range bin widths and verifiable before deployment.

The proposition is a continuous-time statement. Sample-and-hold error, changes in the detection set, and missing detections require additional assumptions and are not covered by $\varepsilon_{\mathrm{meas}}$.
For a stationary obstacle under exact measurements,
\(\widehat{\dot R}_i=-\hat v\cos\hat\gamma_i\) holds.
Substitution into \eqref{eq:cap_robust} cancels the ego-velocity terms. For a positive nominal command, the cap reaches zero at
\begin{equation}
R_{\mathrm{stop}} = R_{\mathrm{safe}} + \frac{\varepsilon_{\mathrm{meas}}}{\alpha}.
\label{eq:standoff}
\end{equation}
This gives the predicted stopping range for a positive nominal forward command under the velocity-input model.
With the values of Table~\ref{tab:params}, \(\varepsilon_{\mathrm{meas}}=0.258\,\mathrm{m/s}\) and \(R_{\mathrm{stop}}=0.758\,\mathrm{m}\). \\

\noindent\textbf{Corollary 1 (Feasible detection range)}
Under the conditions of Proposition~1, a feasible command exists at detection
\(i\) when \(v_{\mathrm{cap},i}\geq -v_{\mathrm{back}}\). For a head-on detection
with \(\hat\gamma_i=0\) and \(\hat v=0\), closing at speed
\(s=-\widehat{\dot R}_i>0\), this holds when
\begin{equation}
\hat R_i \;\geq\; R_{\mathrm{safe}}+\frac{s+\varepsilon_{\mathrm{meas}}-v_{\mathrm{back}}}{\alpha}.
\label{eq:feasible_range}
\end{equation}
With the values of Table~\ref{tab:params}, an obstacle closing at
\(1.0\,\mathrm{m/s}\) must be detected beyond \(1.46\,\mathrm{m}\).

The quantity \(\bar e_{\dot R}\) follows from the measured Doppler bin width, \(\bar e_R\) from the range grid and the calibrated range offset, and \(\bar e_v\) from the robot's own body-velocity estimate. 
The bearing bound is set to \(3^\circ\). We use \(\Delta v_{\max}=v_{\mathrm{des}}+v_{\mathrm{back}} =0.50\,\mathrm{m/s}\), which is the full span of the commanded velocity interval \([-v_{\mathrm{back}},v_{\mathrm{des}}]\). The bound \(\lvert\hat v-v_{\mathrm{cmd}}\rvert\leq\Delta v_{\max}\) therefore assumes that the measured forward velocity used in the constraint remains within this interval. Table~\ref{tab:params} reports each value.

This margin covers bounded errors in range, Doppler, ego velocity, and the bearing projection. It does not cover missed detections, ghost detections, or switching between radar scattering points.

\section{SIMULATION STUDY}
\label{sec:sim}

\subsection{Setup}
\label{sec:setup}

The simulation covers three points. The first is the delay with which each range-rate source reflects a change in obstacle motion. The second is the effect of that delay on safety when little clearance remains. The third is whether added margin or degraded radar detections change the outcome.

Every method uses the same plant, radar model, detection processing, and measurement margin. Only the source of \(\widehat{\dot R}_i\) changes. This comparison isolates the effect of the range-rate source at a common margin.

The plant reproduces the measured Go1 velocity interface. It has a command delay of \(0.15\)--\(0.20\,\mathrm{s}\) and measured acceleration limits. The radar updates at \(20\,\mathrm{Hz}\), and the safety filter runs at \(50\,\mathrm{Hz}\).
The radar model includes the field of view, range gates, missed detections, measurement noise, and quantization at the configured range and Doppler bins. Table~\ref{tab:params} gives the parameters used in all reported cases.

\begin{table}[t]
\caption{Parameters used in simulation and on the robot.}
\label{tab:params}
\centering
\footnotesize
\setlength{\tabcolsep}{4pt}
\begin{tabular}{cc|cc}
\hline
\textbf{Parameter} & \textbf{Value} & \textbf{Parameter} & \textbf{Value}\\
\hline
\(\alpha\) & \(1\,\mathrm{s}^{-1}\) & Radar frequency & \(20\,\mathrm{Hz}\) \\
\(R_{\mathrm{safe}}\) & \(0.50\,\mathrm{m}\) & Range bin & \(0.049\,\mathrm{m}\) \\
\(\bar e_{\dot R}\) & \(0.15\,\mathrm{m/s}\) & Doppler bin & \(0.189\,\mathrm{m/s}\) \\
\(\bar e_R\) & \(0.04\,\mathrm{m}\) & Command delay & \(0.15\)--\(0.20\,\mathrm{s}\) \\
\(\bar e_v\) & \(0.04\,\mathrm{m/s}\) & \(v_{\mathrm{des}}\) & \(0.20\,\mathrm{m/s}\) \\
\(\bar e_\gamma\) & \(3^\circ\) & \(v_{\mathrm{back}}\) & \(0.30\,\mathrm{m/s}\) \\
\(\Delta v_{\max}\) & \(0.50\,\mathrm{m/s}\) & \(R_{\mathrm{stop}}\) & \(0.758\,\mathrm{m}\) \\
\(\varepsilon_{\mathrm{meas}}\) & \(0.258\,\mathrm{m/s}\) & Near boundary reserve & \(0.10\,\mathrm{m}\) \\
\hline
\end{tabular}
\end{table}

We compare direct Doppler with a first difference, the static-obstacle assumption \(\widehat{\dot R}_i=-\hat v\cos\hat\gamma_i\), a constant-velocity Kalman filter (KF-V), and a constant-acceleration Kalman filter (KF-A).

Constant-velocity and constant-acceleration models are also used in an interacting multiple model (IMM) tracker of \cite{mandaokar2026radar}, where the obstacle velocity reaches the barrier through the tracker and not through the Doppler channel. Here, KF-V and KF-A are evaluated separately using range measurements.
All four alternatives use the same quantized range measurements. 
In each Kalman filter, the process-noise value $q$ controls how readily the estimated motion changes when a new range disagrees with the model. A larger value responds faster but passes more range noise. We selected $q=0.1\,\mathrm{m^2/s^3}$ for KF-V and $q=1\,\mathrm{m^2/s^5}$ for KF-A by minimizing the range-rate RMS error against the true rate on a separate calibration set, then fixed them for every simulation and hardware comparison.

The true range rate with the same margin is included as a reference.

The encounter sweep covers initial ranges from \(0.6\) to \(2.0\,\mathrm{m}\) and approach speeds from \(0.2\) to \(1.4\,\mathrm{m/s}\). Each of the 56 combinations uses ten common random seeds, giving 560 paired encounters. 
A reference that commands \(v_{\min}\) from the first frame cannot hold 52 of them. We report these cases as infeasible and exclude them from the method comparison. Of the remaining 508 encounters, 54 leave less than \(0.10\,\mathrm{m}\) of near boundary reserve. These are the near-boundary cases.

A buffer incursion is \(\min_t h<0\), evaluated on the true state with no numerical tolerance. We also report the \(95\)th-percentile incursion depth in the near-boundary set, mean retreat distance, and median response latency. 
Response latency is measured from the first activation of the true robust constraint to the first command change. Every comparison is paired by encounter.

\subsection{Response to a change in motion}
\label{sec:motion}

\begin{figure}[t]
  \centering
  \includegraphics[width=0.95\columnwidth]{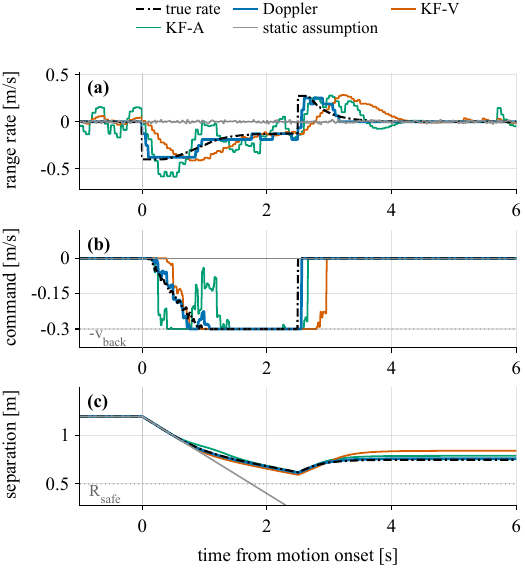}
  \caption{Response to a change in obstacle motion. (a) Range rate supplied to the safety filter. (b) Forward velocity command. (c) True surface separation.}
  \label{fig:motion}
\end{figure}

Fig.~\ref{fig:motion} isolates what happens when the obstacle starts and stops.
The robot is initially stationary at $1.2\,\mathrm{m}$. The obstacle approaches at $0.4\,\mathrm{m/s}$ and stops after moving $1.0\,\mathrm{m}$. Doppler reports the new range rate at the next radar update. A range-based estimator needs several samples to adjust after each change. In this scenario, the Doppler command changes $0.09\,\mathrm{s}$ after the constraint becomes active, compared with $0.33\,\mathrm{s}$ for KF-V. 
KF-A changes at the same first sample as Doppler in this case, but continues to report motion after the obstacle stops. During the approach, the Doppler range-rate RMS error is $0.059\,\mathrm{m/s}$, about one third of the two Kalman-filter errors.

The margin must cover the range-rate error while the constraint is active. Doppler remains inside $\bar e_{\dot R}=0.15\,\mathrm{m/s}$ in the tested motion profile. Both Kalman filters exceed this bound in the approach case while their estimates adjust to the new motion.

\subsection{Paired encounter sweep}
\label{sec:sweep}

The motion-change test gives the mechanism behind the different
reactions. The paired sweep applies the same encounter to every method and compares the closed-loop safety result. Table~\ref{tab:results} gives the closed-loop result. Direct Doppler has two incursions in 508 evaluable encounters and one in the 54 near-boundary encounters. 
KF-A and KF-V have 15 and 25 incursions in total. Their mean retreat distances are on opposite sides of the Doppler value. The difference in safety is therefore not explained by one method always retreating farther.

\begin{table}[t]
\caption{Paired sweep over the 508 evaluable encounters. Bold entries mark the best deployable safety and response values. Retreat is the cost of that safety and has no preferred direction.}
\label{tab:results}
\centering
\begin{tabular}{cccccc}
\hline
\textbf{Range rate} & \textbf{Incursions} & \textbf{Near} & \textbf{p95 depth} & \textbf{Retreat} & \textbf{Latency} \\
\textbf{source} & \textbf{/508} & \textbf{/54} & \textbf{[m]} & \textbf{[m]} & \textbf{[s]} \\
\hline
True \(\dot R\)& 0 & 0 & 0.000 & 0.213 & 0.00 \\
\textbf{Doppler} & \textbf{2} & \textbf{1} & \textbf{0.000} & 0.233 & \textbf{0.00} \\
KF-A & 15 & 15 & 0.027 & 0.210 & 0.06 \\
KF-V & 25 & 16 & 0.042 & 0.280 & 0.06 \\
\(\Delta R/\Delta t\) & 98 & 21 & 0.076 & 0.126 & 0.06 \\
Static & 280 & 40 & 0.304 & 0.000 & 0.60 \\
\hline
\end{tabular}
\end{table}

\begin{figure}[t]
  \centering
  \includegraphics[width=0.98\columnwidth]{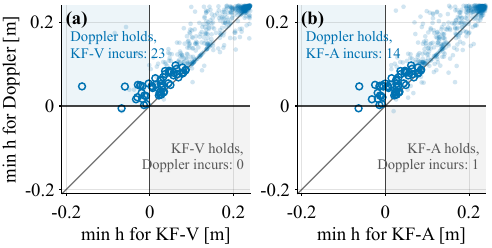}
  \caption{Minimum true barrier value in the same encounter under Doppler and under (a) KF-V or (b) KF-A. Open circles are near-boundary encounters. The shaded quadrant contains cases held by Doppler and lost by the range-rate estimator, and vice versa.}
  \label{fig:paired}
\end{figure}

The paired comparison in Fig.~\ref{fig:paired} shows where the difference occurs. Against KF-V, Doppler holds 23 encounters that KF-V loses, with no encounter in the opposite quadrant (exact McNemar test, \(p<10^{-6}\)). 
Against KF-A, the corresponding counts are 14 and 1. 

\subsection{Margin and radar failures}
\label{sec:robust}

The final simulation test covers two items: the margin required to compensate for a range-rate estimator, and the response to the two radar failures observed on the robot.
With no measurement margin, the Doppler controller aims at the boundary and incurs in all 508 evaluable encounters. The sensor-error margin reduces this to two. 

\begin{figure}[t]
  \centering
  \includegraphics[width=0.98\columnwidth]{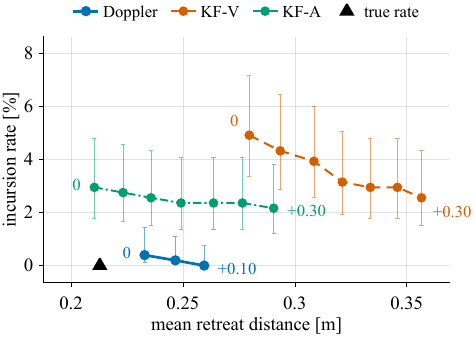}
  \caption{Incursion rate over the 508 encounters against mean retreat as margin is added to each range-rate source. Labels give the added margin in \(\mathrm{m/s}\). The triangle is the true range rate with the same operating margin. Vertical bars are $95\,\%$ Wilson intervals for the incursion rate.}
  \label{fig:margin}
\end{figure}

Fig.~\ref{fig:margin} adds a constant margin to each range-rate source. At an added \(0.30\,\mathrm{m/s}\), KF-V and KF-A still incur in 11 of the 54 near-boundary encounters. Doppler incurs in one at the nominal margin and none after adding \(0.10\,\mathrm{m/s}\).

For the tested estimator settings, increasing the margin reduces incursions but leaves a gap to Doppler within the tested margin range. Fig.~\ref{fig:margin} also shows the associated retreat cost. 

We also apply the two radar failures seen during the hardware study as controlled simulation conditions. At a \(20\,\%\) per-frame scatterer-switching rate, the incursion counts are 7 for Doppler, 43 for KF-V, and 79 for KF-A. 
With detection probability reduced to \(0.60\), the counts are 7 and 101 for Doppler and KF-V. A range jump enters the range-rate estimators as a large velocity. Doppler is unchanged by that jump, although the range term in the constraint still changes for one frame. Missing detections age the direct measurement and also add prediction error to a temporal estimator.

The guarantee still requires feasibility. If an obstacle keeps approaching faster than the robot can retreat, no command in \([v_{\min},v_{\max}]\) can hold the buffer. The 52 infeasible sweep cases are kept separate for this reason.

\section{HARDWARE EXPERIMENTS}
\label{sec:hw}

\subsection{Platform and processing}

The hardware study tests the stopping range predicted by \eqref{eq:standoff} and the selection of the most restrictive return in a two-object encounter. The range-rate sources are then compared through command timing and stationary retreat activity.

The experiments use a Unitree Go1 with a Texas Instruments IWR6843AOPEVM FMCW radar mounted level on its back. The radar operates from \(60\) to \(64\,\mathrm{GHz}\) and produces a point cloud at \(20\,\mathrm{Hz}\). 
The deployed configuration gives a \(0.049\,\mathrm{m}\) range bin and a \(0.189\,\mathrm{m/s}\) Doppler bin. An Intel RealSense D435 above the radar records metric depth for the range-sensor baseline. 

Radar points are gated by SNR, height, range, bearing, and a \(\lvert y\rvert\leq0.5\,\mathrm{m}\) forward corridor. Points within \(0.30\,\mathrm{m}\) are grouped. The member with the smallest range supplies the group range and bearing. 
Group Doppler is an outlier-rejected mean over the member returns. Groups are associated one-to-one within \(0.40\,\mathrm{m}\) and kept
for at most eight radar frames when missed. This association maintains group
continuity but does not estimate obstacle velocity, and the controller does
not differentiate range. Each group enters \eqref{eq:cap_robust} as one detection, and \eqref{eq:scalar_safety_filter} selects the smallest cap. Any mismatch between the selected member's range rate and the group Doppler is outside the measurement-error margin in \eqref{eq:eps_meas}.

The depth baseline converts each image to a planar scan with one nearest point per image column in the selected height band. The scan points are grouped with a \(0.40\,\mathrm{m}\) radius and use the same one-to-one association. 
First difference, KF-V, and KF-A then supply the missing range rate.
The depth camera tests whether similar command delays and stationary retreats occur with another range-only stream.

\begin{figure}[t]
  \centering
  \includegraphics[width=0.98\columnwidth]{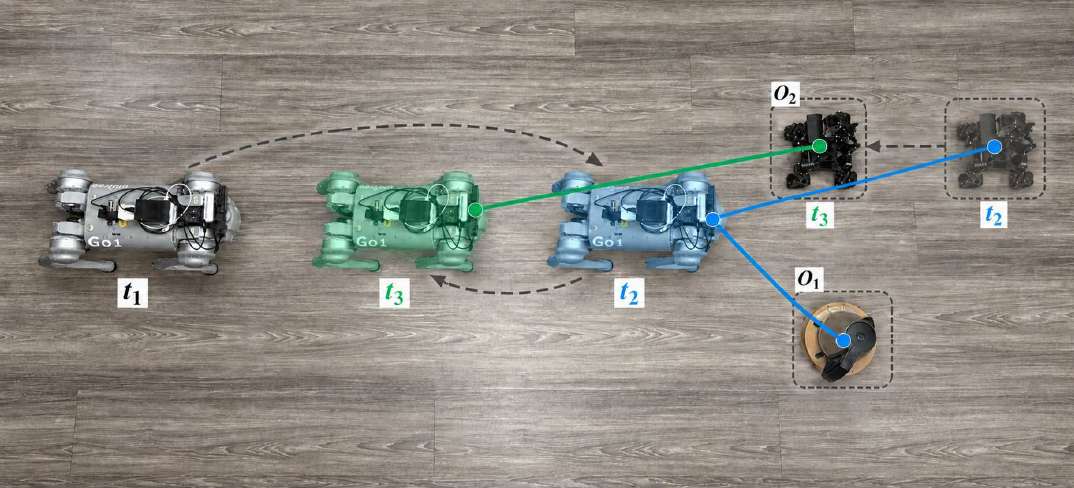}
  \caption{Composite illustration of the hardware and test sequence. The robot first approaches the static object $O_1$ and stops. A second object $O_2$ then approaches from farther away. The reported dynamic trials use a person or a moving robot.}
  \label{fig:hardware}
\end{figure}

The static scene tests the static standoff. The robot walks toward a static obstacle $O_1$ at \(0.20\,\mathrm{m/s}\) and stops under the filtered command. 
The dynamic scene tests multi-object selection (Fig.~\ref{fig:hardware}). The robot first stops at the static obstacle $O_1$. A person or remote-controlled small robot ($O_2$) then approaches from farther away, stops near the static obstacle, and walks back. 
We recorded 23 evaluated radar approaches and 9 with synchronized depth.

\subsection{Static standoff}

The parked range is the offset-corrected radar range to $O_1$. The $10$ static trials park between $0.760$ and $0.800\,\mathrm{m}$ stopping distance, with mean $0.777 \pm 0.015\,\mathrm{m}$, against the predicted $R_{\mathrm{stop}} = 0.758\,\mathrm{m}$.
The mean difference is $0.019\,\mathrm{m}$, and every trial is within one range bin above the prediction. This tests the measurable consequence of the margin in
\eqref{eq:standoff}.

\subsection{Response to a farther approaching object}

\begin{figure}[t]
  \centering
  \includegraphics[width=0.98\columnwidth]{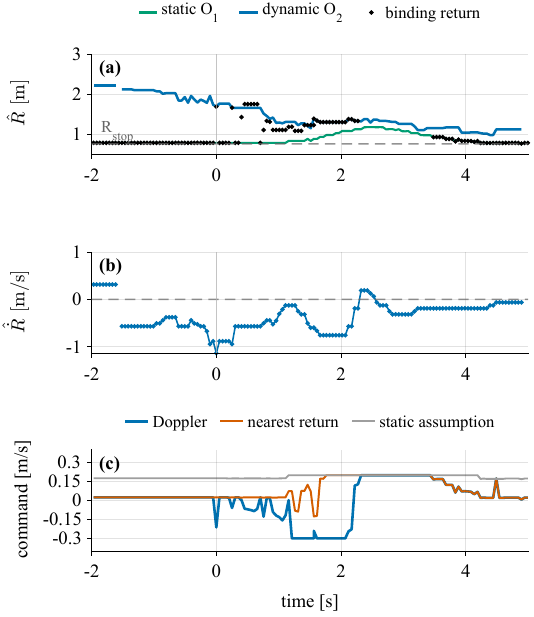}
  \caption{Representative two-object pass. (a) Object ranges and the return that sets the minimum cap. (b) Doppler range rate of the approaching-object group. (c) Doppler, nearest-return, and static-assumption commands.}
  \label{fig:hw_dyn}
\end{figure}

The two-object test separates nearest range from range rate.
Fig.~\ref{fig:hw_dyn} shows a pass in which the nearest return is not the binding one. The robot stops at the static object $O_1$ at \(0.783\,\mathrm{m}\), half a range bin outside \(R_{\mathrm{stop}}\), when the approaching object $O_2$ closes from \(1.69\,\mathrm{m}\). 
The static object admits \(+0.025\,\mathrm{m/s}\) and the approaching object \(-0.213\,\mathrm{m/s}\), so the farther return sets the cap. The onboard command follows one frame later.
The nearest-return baseline binds to the nearest static object and retreats at \(1.26\,\mathrm{s}\). The static-obstacle assumption does not retreat at all, as it does not consider relative velocities.

\subsection{Range-rate source comparison}

The aggregate comparison tests two behaviors together: starting a retreat command during an approach and avoiding retreat commands while nothing is closing. Table~\ref{tab:hw_sources} reports the exact response counts, the lag vs. Doppler, and stationary retreats.

\begin{table}[t]
\caption{Hardware comparison of range-rate sources. Doppler defines the reference sets for response and retreat activity. Lag is the median delay of the first retreat command relative to Doppler; positive values indicate later responses.}
\label{tab:hw_sources}
\centering
\begin{tabular}{cccc}
\hline
\textbf{Range-rate}  & \textbf{Response} & \textbf{Lag vs.} & \textbf{Stationary retreat} \\
            \textbf{source}      & \textbf{[count]}  & \textbf{Doppler [s]}         & \textbf{[\% frames]} \\
\hline
Radar KF-V
    & 22/23 & +0.15 & 5.3 \\
Radar KF-A
    & 22/23 & +0.08 & 8.6 \\
Radar $\Delta R/\Delta t$
    & 23/23 & +0.25 & 23.6 \\
Depth KF-V
    & 9/9 & +0.25 & 5.7 \\
Depth KF-A
    & 9/9 & +0.20 & 7.1 \\
Depth $\Delta R/\Delta t$
    & 9/9 & +0.20 & 19.1 \\
\hline
\end{tabular}
\end{table}

On radar, the Kalman filters respond in 22 of the 23 approaches, with median delays of \(0.08\)--\(0.15\,\mathrm{s}\) relative to Doppler. The first difference responds in all 23 approaches, but retreats in $23.6\,\%$ of stationary frames.
The depth camera checks whether the same behavior appears with another range sensor. It has lower range noise, the lags grow to $0.20$ to $0.25\,\mathrm{s}$ and the stationary retreat rates stay close to the radar values. 
Reducing range noise does not remove the lag and every range-derived source responds after Doppler on both sensors.

\section{CONCLUSION}

Direct Doppler provides the derivative needed by an ego-centered range barrier without an estimated obstacle-velocity state. In the tested motion profiles, the range-based Kalman filters can exceed the selected range-rate error bound after a change in obstacle motion. For the reported filter settings, increasing the fixed margin reduces incursions but leaves near-boundary failures within the tested margin range.
Doppler's lower incursion count is not caused by uniformly stronger retreat, as its mean retreat distance lies between KF-A and KF-V. 

On the robot, the controller settles within one range bin of the predicted static standoff and a farther closing object can restrict the command while a nearer static return remains. The hardware comparison reports command timing and retreat activity relative to Doppler.

The result is limited to the longitudinal range channel and to obstacles returned by the radar. The forward-invariance claim further requires bounded measurement errors, continuously available detections, continuous constraint enforcement, exact command tracking, and a feasible command. 
The reported distance refers to a radar scattering point, the group Doppler is a mean over member returns rather than a single detection, and the forward corridor can lose an object after lateral drift. Extending the guarantee to sampled-data updates and changing detection sets, covering the tangential channel, and coupling the range constraint with steering are the main next steps.






\bibliographystyle{IEEEtran}
\bibliography{references}

@article{ames2019control,
  author  = {Ames, Aaron D. and Coogan, Samuel and Egerstedt, Magnus
             and Notomista, Gennaro and Sreenath, Koushil
             and Tabuada, Paulo},
  title   = {Control Barrier Functions: Theory and Applications},
  journal = {Proc. European Control Conf.},
  year    = {2019},
  pages   = {3420--3431}
}

@inproceedings{todorovski2026modular,
  title={Modular design of strict control {Lyapunov} functions for global stabilization of the unicycle in polar coordinates},
  author={Todorovski, Velimir and Kim, Kwang Hak and Krsti{\'c}, Miroslav},
  booktitle={2026 American Control Conference (ACC)},
  pages={1122--1127},
  year={2026},
  organization={IEEE}
}

@article{harlow2024new,
  title={A new wave in robotics: Survey on recent {mmWave} radar applications in robotics},
  author={Harlow, Kyle and Jang, Hyesu and Barfoot, Timothy D and Kim, Ayoung and Heckman, Christoffer},
  journal={IEEE Transactions on Robotics},
  volume={40},
  pages={4544--4560},
  year={2024},
  publisher={IEEE}
}

@inproceedings{dean2021guaranteeing,
  title={Guaranteeing safety of learned perception modules via measurement-robust control barrier functions},
  author={Dean, Sarah and Taylor, Andrew and Cosner, Ryan and Recht, Benjamin and Ames, Aaron},
  booktitle={Conference on Robot Learning},
  pages={654--670},
  year={2021},
  organization={PMLR}
}

@inproceedings{cosner2021measurement,
  title={Measurement-robust control barrier functions: Certainty in safety with uncertainty in state},
  author={Cosner, Ryan K and Singletary, Andrew W and Taylor, Andrew J and Molnar, Tamas G and Bouman, Katherine L and Ames, Aaron D},
  booktitle={2021 IEEE/RSJ International Conference on Intelligent Robots and Systems (IROS)},
  pages={6286--6291},
  year={2021},
  organization={IEEE}
}

@inproceedings{yang2023safe,
  title={Safe perception-based control under stochastic sensor uncertainty using conformal prediction},
  author={Yang, Shuo and Pappas, George J and Mangharam, Rahul and Lindemann, Lars},
  booktitle={2023 62nd IEEE Conference on Decision and Control (CDC)},
  pages={6072--6078},
  year={2023},
  organization={IEEE}
}

@article{long2026sensor,
  title={Sensor-based distributionally robust control for safe robot navigation in dynamic environments},
  author={Long, Kehan and Yi, Yinzhuang and Dai, Zhirui and Herbert, Sylvia and Cort{\'e}s, Jorge and Atanasov, Nikolay},
  journal={The International Journal of Robotics Research},
  volume={45},
  number={2},
  pages={328--351},
  year={2026},
  publisher={SAGE Publications Sage UK: London, England}
}

@inproceedings{wang2022observer,
  title={Observer-based control barrier functions for safety critical systems},
  author={Wang, Yujie and Xu, Xiangru},
  booktitle={2022 American Control Conference (ACC)},
  pages={709--714},
  year={2022},
  organization={IEEE}
}

@inproceedings{ames2014control,
  title={Control barrier function based quadratic programs with application to adaptive cruise control},
  author={Ames, Aaron D and Grizzle, Jessy W and Tabuada, Paulo},
  booktitle={53rd IEEE Conference on Decision and Control},
  pages={6271--6278},
  year={2014},
  organization={IEEE}
}

@article{dai2024sailing,
  title={Sailing through point clouds: Safe navigation using point cloud based control barrier functions},
  author={Dai, Bolun and Khorrambakht, Rooholla and Krishnamurthy, Prashanth and Khorrami, Farshad},
  journal={IEEE Robotics and Automation Letters},
  volume={9},
  number={9},
  pages={7731--7738},
  year={2024},
  publisher={IEEE}
}

@article{keyumarsi2023lidar,
  title={{LiDAR}-based online control barrier function synthesis for safe navigation in unknown environments},
  author={Keyumarsi, Shaghayegh and Atman, Made Widhi Surya and Gusrialdi, Azwirman},
  journal={IEEE Robotics and Automation Letters},
  volume={9},
  number={2},
  pages={1043--1050},
  year={2023},
  publisher={IEEE}
}

@article{raja2024ogm,
  title={{OGM-CBF}: Occupancy Grid Map-based Control Barrier Function for Safe Mobile Robot Control with Memory of out of View Obstacles},
  author={Raja, Golnaz and Pr{\'a}gr, Milo{\v{s}} and K{\"a}rki, Topi Reino Johannes and M{\"o}kk{\"o}nen, Teemu and Ghabcheloo, Reza},
  journal={arXiv preprint arXiv:2405.10703},
  year={2024}
}

@article{zhang2026collision,
  title={A collision-free reactive guidance module for safety-critical vehicles using local measurements},
  author={Zhang, Fei and Yang, Guang-Hong},
  journal={IEEE Transactions on Systems, Man, and Cybernetics: Systems},
  year={2026},
  publisher={IEEE}
}

@article{molnar2021model,
  title={Model-free safety-critical control for robotic systems},
  author={Molnar, Tamas G and Cosner, Ryan K and Singletary, Andrew W and Ubellacker, Wyatt and Ames, Aaron D},
  journal={IEEE Robotics and Automation Letters},
  volume={7},
  number={2},
  pages={944--951},
  year={2021},
  publisher={IEEE}
}

@article{oruganti2023robust,
  title={Robust control barrier functions for sampled-data systems},
  author={Oruganti, Pradeep Sharma and Naghizadeh, Parinaz and Ahmed, Qadeer},
  journal={IEEE Control Systems Letters},
  volume={8},
  pages={103--108},
  year={2023},
  publisher={IEEE}
}

@article{tayal2026collision,
  title={A collision cone approach for control barrier functions},
  author={Tayal, Manan and Goswami, Bhavya Giri and Rajgopal, Karthik and Singh, Rajpal and Rao, M Tejas and Keshavan, Jishnu and Jagtap, Pushpak and Yadukumar, Shishir Nadubettu},
  journal={IEEE Transactions on Control Systems Technology},
  year={2026},
  publisher={IEEE}
}

@article{bilik2019rise,
  title={The rise of radar for autonomous vehicles: Signal processing solutions and future research directions},
  author={Bilik, Igal and Longman, Oren and Villeval, Shahar and Tabrikian, Joseph},
  journal={IEEE Signal Processing Magazine},
  volume={36},
  number={5},
  pages={20--31},
  year={2019},
  publisher={IEEE}
}

@inproceedings{kramer2020radar,
  title={Radar-inertial ego-velocity estimation for visually degraded environments},
  author={Kramer, Andrew and Stahoviak, Carl and Santamaria-Navarro, Angel and Agha-Mohammadi, Ali-Akbar and Heckman, Christoffer},
  booktitle={2020 IEEE International Conference on Robotics and Automation (ICRA)},
  pages={5739--5746},
  year={2020},
  organization={IEEE}
}

@article{singletary2021safety,
  title={Safety-critical kinematic control of robotic systems},
  author={Singletary, Andrew and Kolathaya, Shishir and Ames, Aaron D},
  journal={IEEE Control Systems Letters},
  volume={6},
  pages={139--144},
  year={2021},
  publisher={IEEE}
}

@article{bruhl2024making,
  title={Making Radar Detections Safe for Autonomous Driving: A Review},
  author={Br{\"u}hl, Tim and Ewecker, Lukas and Schwager, Robin and Sohn, Tin Stribor and Hohmann, S{\"o}ren and Vinel, AV and Berns, K and Ploeg, J and Gusikhin, O},
  journal={VEHITS},
  pages={299--310},
  year={2024}
}

@inproceedings{bijelic2020seeing,
  title={Seeing through fog without seeing fog: Deep multimodal sensor fusion in unseen adverse weather},
  author={Bijelic, Mario and Gruber, Tobias and Mannan, Fahim and Kraus, Florian and Ritter, Werner and Dietmayer, Klaus and Heide, Felix},
  booktitle={2020 IEEE/CVF Conference on Computer Vision and Pattern Recognition (CVPR)},
  pages={11679--11689},
  year={2020},
  organization={IEEE}
}

@article{mandaokar2026radar,
  title={Radar Perception for Dynamic Obstacle Avoidance onboard small-scale Quadrotor {UAVs}},
  author={Mandaokar, Dnyandeep and Rinner, Bernhard},
  journal={arXiv preprint arXiv:2608.01855},
  year={2026}
}

@inproceedings{autenrieb2025sensor,
  title={Sensor-based safety-critical control using an incremental control barrier function formulation via reduced-order approximate models},
  author={Autenrieb, Johannes and Shin, Hyo-Sang},
  booktitle={2025 American Control Conference (ACC)},
  pages={374--381},
  year={2025},
  organization={IEEE}
}

\end{document}